\documentclass{article}

\usepackage{arxiv}

\usepackage[utf8]{inputenc} 
\usepackage[T1]{fontenc}    
\usepackage{hyperref}       
\usepackage{url}            
\usepackage{booktabs}       
\usepackage{amsfonts}       
\usepackage{nicefrac}       
\usepackage{microtype}      
\usepackage{fancyhdr}       
\usepackage{graphicx}       
\graphicspath{{media/}}     

\title{The Sociolinguistic Foundations \\ of Language Modeling}

\author{
  Jack Grieve
  \\j.grieve@bham.ac.uk
  \And Sara Bartl
  \\s.bartl@bham.ac.uk
  \And Matteo Fuoli
  \\m.fuoli@bham.ac.uk
  \And Jason Grafmiller
  \\j.grafmiller@bham.ac.uk
  \And Weihang Huang
  \\w.huang.5@bham.ac.uk
  \And Alejandro Jawerbaum
  \\a.napolitanojawerbaum@bham.ac.uk
  \And  Akira Murakami
  \\a.murakami@bham.ac.uk
  \And Marcus Perlman
  \\m.perlman@bham.ac.uk
  \And Dana Roemling
  \\d.roemling@bham.ac.uk
  \And Bodo Winter
  \\b.winter@bham.ac.uk
  \And
  \\
  Department of Linguistics and Communication \\
  University of Birmingham
}

\begin{document}
\maketitle
\begin{abstract}
In this paper, we introduce a sociolinguistic perspective on language modeling. We claim that large language models are inherently models of \textit{varieties of language}, and we consider how this insight can inform the development and deployment of large language models. We begin by presenting a technical definition of the concept of a variety of language as developed in sociolinguistics. We then discuss how this perspective can help address five basic challenges in language modeling: \textit{social bias}, \textit{domain adaptation}, \textit{alignment}, \textit{language change}, and \textit{scale}. Ultimately, we argue that it is crucial to carefully define and compile training corpora that accurately represent the specific varieties of language being modeled to maximize the performance and societal value of large language models.
\end{abstract}


\section{Introduction}

The underlying task of language modeling is to predict the probability of a word, or other linguistic forms, in a text based on previously observed texts \cite{JurafskyMartin2023}. Language modeling is not new \cite{BengioDucharmeVincentJauvin2003}, but when pursued through the analysis of extremely large corpora of natural language using transformer-based architectures \cite{VaswaniShazeerParmarUszkoreitJonesGomez2017,DevlinChangLeeToutanova2018}, it has proven to be a uniquely effective approach to natural language processing (NLP) \cite{RadfordWuChildLuanAmodeiSutskever2019}. These systems, which have come to be known as Large Language Models (LLMs), are currently revolutionizing Artificial Intelligence (AI), with especially powerful LLMs like GPT-4 \cite{AchiamAdlerAgarwalAhmadAkkayaAleman} and LLaMa \cite{TouvronMartinStoneAlbertAlmahairiBabaei2023} often being referred to as base models or foundation models \cite{BommasaniHudsonAdeliAltmanAroravonArx2021} due to their high levels of fluency and their ability to help achieve state-of-the-art performance across a wide range of downstream tasks, most famously in chatbots like ChatGPT \cite{Ray2023}. Despite increasing concerns about the risks of LLMs \cite{BenderGebruMcMillanMajorShmitchell2021}, experts across many fields believe they will have a major impact on society, including in medicine \cite{ThirunavukarasuTingElangovanGutierrezTanTing2023}, education \cite{KasneciSesslerKuchemannBannertDementievaFischer2023}, computer programing \cite{LiChoiChungKushmanSchrittwieserLeblond2022}, journalism \cite{Pavlik2023}, and technical writing \cite{LundWangMannuruNieShimrayWang2023}. 

Given the growing societal importance of LLMs, language modeling has provoked critical discussion from a wide range of perspectives, not only AI and NLP (e.g., \cite{BenderGebruMcMillanMajorShmitchell2021,BommasaniHudsonAdeliAltmanAroravonArx2021}), but in linguistics (e.g., \cite{Piantadosi2023,DentellaGuntherLeivada2023,MarcusLeivadaMurphy2023}), cognitive science (e.g., \cite{HardySucholutskyThompsonGriffiths2023,DemszkyYangYeagerBryanClapperChandhok2023,MichaelovBardolphVanPettenBergenCoulson2024}, and ethics (e.g., \cite{BirhaneKasirzadehLeslieWachter2023,CabreraLoyolaMaganaRojas2023,LiMoonPurkayasthaCeliTrivediGichoya2023,StefanCarutasuMocan2023,HaqueLi2024}). There is, however, a very basic question about language modeling that has received remarkably little attention in the literature: 

\begin{quote}
What is actually being modeled by language models?
\end{quote}

Although the goal of language modeling is clear (i.e. token prediction), the type of language being modeled by language models is only ever defined in the most general terms, for example, “a broad swath of internet data” \cite{BrownMannRyderSubbiahKaplanDhariwal2020}. Models are often trained on corpora based at least in part on the CommonCrawl dataset \cite{RadfordWuChildLuanAmodeiSutskever2019,RaffelShazeerRobertsLeeNarangMatena2020,Baack2024}, but otherwise, in most cases, the nature of the language being modeled is not described at all \cite{BenderGebruMcMillanMajorShmitchell2021}. In large part, this is a natural consequence of the need for massive amounts of data to train base models, making the sources of these corpora of secondary concern. However, even when these models are adapted for more specific contexts \cite{GururanganMarasovicSwayamdiptaLoBeltagyDowneySmith2020}, the type of language used for further training is generally only loosely defined. For example, ChatGPT was developed by adapting a GPT-3.5 base model for dialogue \cite{noauthor_chatgpt_nodate}, but the form of dialogue actually being modeled by ChatGPT is something much less diverse and much more artificial than everyday English conversation, as anyone who interacts with ChatGPT knows. 

Drawing on modern sociolinguistic theory, in this paper, we therefore provide an answer to the question what is being modeled by language models?
\begin{quote}
Language models are models of \textbf{varieties of language}.
\end{quote}
We argue that any language model is inherently modeling the variety of language represented by the corpus on which it is trained, even if that variety of language is unknown and even if that corpus is a poor representation of that variety of language. Our view is that this simple insight can inform, at a fundamental level, how language models are developed and deployed in the real world. Given rapid advances in language modeling in recent years and the increasing societal impact and risk associated with LLMs, we believe the sociolinguistic perspective we are advocating for in this paper is especially important at this time – not only to improve the performance, evaluation, and applicability of LLMs, but to guide the creation of safe and ethical AI systems and to help us better understand their underlying nature.

In the rest of this paper, we expand on our basic claim that language models represent varieties of language and consider the implications of this claim for the future of language modeling. We first provide a technical definition of the sociolinguistic concept of a variety of language and argue that this concept inherently underpins the task of language modeling. We then introduce and discuss five general challenges in language modeling that we believe the sociolinguistic perspective introduced in this paper can help address. We refer to these challenges as \textit{social bias}, \textit{domain adaptation}, \textit{alignment}, \textit{language change}, and \textit{scale}. Our core message is that to maximize the value of LLMs in society, it is crucial to carefully consider the specific varieties of language being modeled and to compile corpora that accurately represent these varieties of language, grounded in theories and methods developed in sociolinguistics for understanding language variation and change.  

\section{Defining Varieties of Language}

A \textit{variety of language}, or more simply a \textit{variety}, is a term commonly used across linguistics to refer to any type of language \cite{CrystalDavy1969,HartmannStork1972,Matthews1997,McEneryXiaoTono2006,Jackson2007,Crystal2011}. The term is especially common in fields that study language variation and change – like sociolinguistics, dialectology, typology, historical linguistics, discourse analysis, stylistics, and corpus linguistics – where it is generally used to identify the types of language targeted for description, comparison, or other forms of linguistic analysis. 

One reason a variety of language is such a powerful concept is because it can be used to identify a wide range of phenomena – from very broadly defined varieties like the entire English language to very narrowly defined varieties like the speeches of a single politician. This terminology also allows linguists to sidestep debates, which are often underlyingly political in nature, like whether a given variety qualifies as a dialect or a language \cite{Meyerhoff2018}. For example, regardless of whether Scots is considered to be a dialect of English or a distinct language, Scots can be considered to be a variety, as well as a sub-variety of some larger Anglic variety that also includes English \cite{Aitken1985}. Similarly, regardless of whether Chinese is considered to be a family composed of many languages or a language composed of many dialects, all forms of Chinese can be considered to be both varieties themselves and part of some larger Sinitic variety \cite{HuangGrieveJiaoCai2024}.

\begin{figure}
    \centering
    \includegraphics[width=1\linewidth]{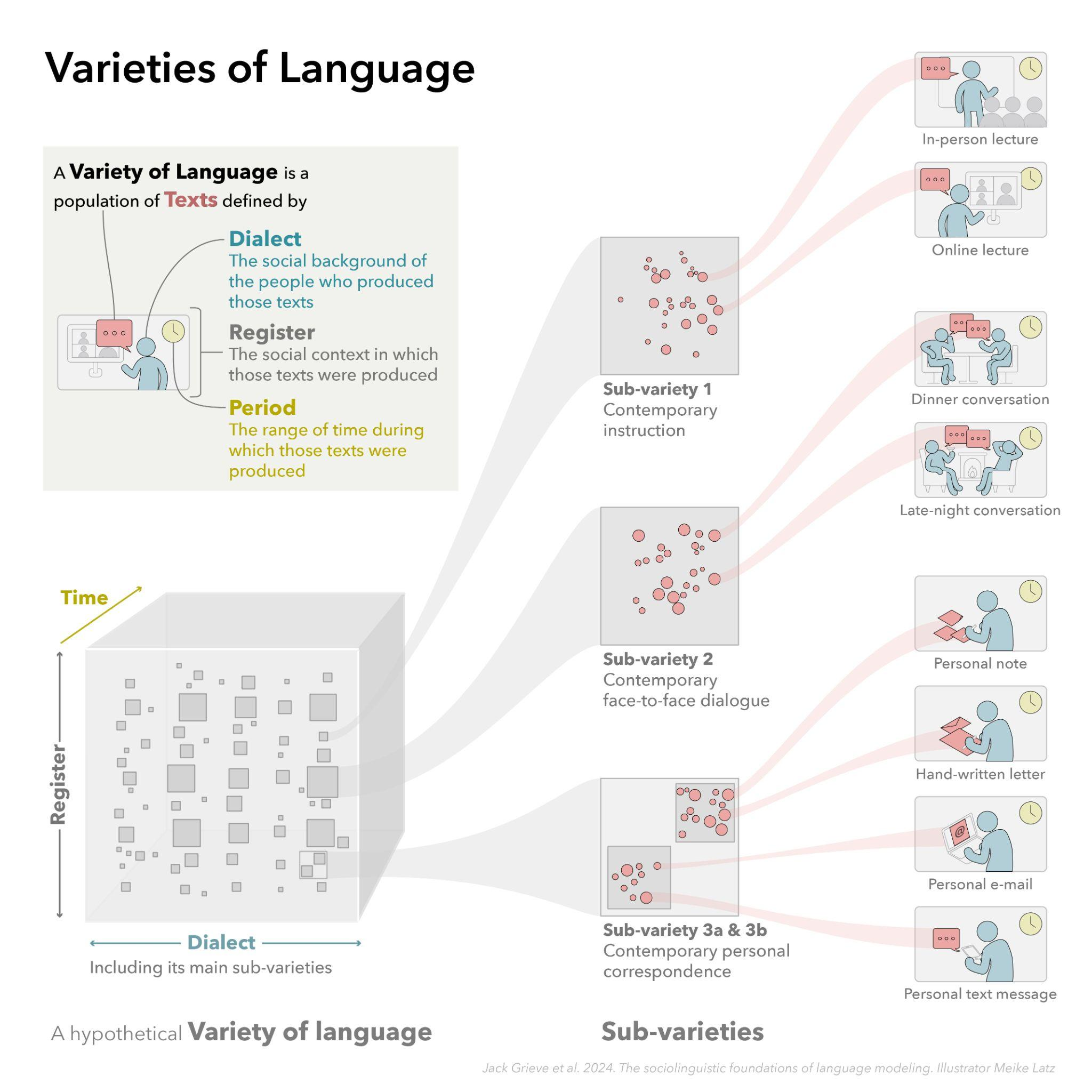}
    \caption{\textit{Varieties of Language}. This figure defines the concept of a variety of language, illustrating how the interaction between three distinct extra-linguistic factors – the social background of people who produce language (dialect), the social context in which language is produced (register), and the range of time over which language is produce (period) – can be used to specify a variety of language. It also illustrates how varieties of language are hierarchically organized, composed of smaller and smaller sub-varieties.}
    \label{fig:fig1varietiesoflanguage}
\end{figure}

Although what are traditionally considered entire languages like English or Chinese can be referred to as varieties, the term is most commonly used in linguistics to refer to more narrowly defined types of language \cite{Crystal2011,Meyerhoff2018,WardhaughFuller2021}. Such varieties are referred to by a wide range of technical and colloquial terms, including not only \textit{dialects}, but \textit{accents}, \textit{sociolects}, \textit{topolects}, \textit{argots}, \textit{jargons}, \textit{registers}, \textit{genres}, \textit{styles}, \textit{slangs}, \textit{standards}, \textit{periods}, and \textit{eras}. We believe, however, that it is especially insightful to recognise three basic and distinct types of varieties – or, alternatively, three basic and distinct sources of linguistic variation – which we refer to as dialect, register, and period (see Figure~\ref{fig:fig1varietiesoflanguage}).

\textbf{Dialects} are varieties defined by the social backgrounds and identities of the people who produce language \cite{ChambersTrudgill1998}. For example, dialects are often associated with language that originates from speakers from a particular nation, region, class, or ethnicity. Crucially, dialects are defined by the social characteristics of language users. Alternatively, \textbf{registers} are varieties defined by the social contexts in which people, potentially from any social background, produce language \cite{BiberConrad2019}. For example, registers are often associated with language produced in specific modalities, media, settings, and topics. Whereas dialects are defined by the characteristics of language users, registers are defined by the characteristics of the contexts in which those language users communicate. Finally, \textbf{periods} are varieties defined by the time span over which language is produced \cite{NevalainenRaumolinBrunberg2016}. Taken together, these three extra-linguistic sources of linguistic variation allow for varieties of language to be defined with great flexibility.

The relationships between varieties can be highly complex (see Figure~\ref{fig:fig1varietiesoflanguage}). Varieties can be defined at any scale and are generally hierarchically structured, being divisible into smaller and smaller sub-varieties. For example, English is a variety, but it also contains many smaller sub-varieties. These include many dialects, including national varieties of English, like British and American English, which are themselves composed of many smaller regional dialects \cite{ChambersTrudgill1998}. At the most narrowly defined level, the language of an individual can even be considered a distinct dialect (i.e., an idiolect). Similarly, English also includes many registers, including spoken and written English, which are themselves composed of many smaller registers, like conversations, telephone conversations, and personal telephone conversations \cite{BiberConrad2019}.

\clearpage
\hspace*{-2in}
\clearpage

Along with exhibiting hierarchical structure, varieties can also be defined based on the overlap of larger varieties (see Figure~\ref{fig:fig1varietiesoflanguage}). For example, it is common to define a variety of interest by specifying a dialect, register, and period, like \textit{Contemporary Conversational Canadian French} or \textit{Scottish Novels from the Twentieth Century Written by Women}. In other words, we can think of a variety as being defined by the specification of one or more extra-linguistic factors related to the circumstances in which language is produced. In addition, the boundaries between varieties are not necessarily sharp or fixed. For example, one regional dialect or literary register might transition gradually into the next, and where we draw a line between them may change over time. 

Although we have defined a variety of language as a type of language, it is important to specify what exactly a variety of language consists of. In other words, when linguists study a variety of language, what are they actually studying? For many linguists, a variety of language is essentially a population of texts (or utterances), as circumscribed by one or more extra-linguistic factors, in particular, by a specific dialect, register, and period (for related discussion, see \cite{Croft2000}). Notably, in this case, a \textbf{text} is broadly defined as the language (e.g., utterances, discourse) produced during any communicative event, including crucially language produced in any modality (e.g., speech, writing, signing) \cite{HallidayHasan1976}. For example, not only can an email or an essay be considered a text, but so can a conversation or a speech. If we adopt what is known as an externalist approach to linguistics \cite{ScholzPelletierPullumNefdt2024,Sampson2002}, where language in general is defined as the population of all texts (or utterances) that have ever been produced, a variety of language can then be defined as a sub-population of those texts that meets some external definition – i.e., the totality of language produced by people from a particular social background (dialect), in a particular social context (register), and over a particular period of time (period).

For example, Contemporary Spoken French Canadian Conversation can be considered a variety of language, as it is a population of texts (i.e., conversations) produced by individuals from a specific social background (i.e., people who live in Canada), in a specific social context (i.e., spoken interactions), during a specific period (i.e., now). Similarly, a more narrowly defined type of language like Scottish Novels from the Twentieth Century Written by Women can also be considered a variety of language, as it is a population of texts (i.e., books) produced by individuals from a specific social background (i.e., female authors from Scotland), in a specific social context (i.e., long-form fictional narratives), during a specific time span (i.e., 1900-1999). 

\begin{figure}
    \centering
    \includegraphics[width=\linewidth]{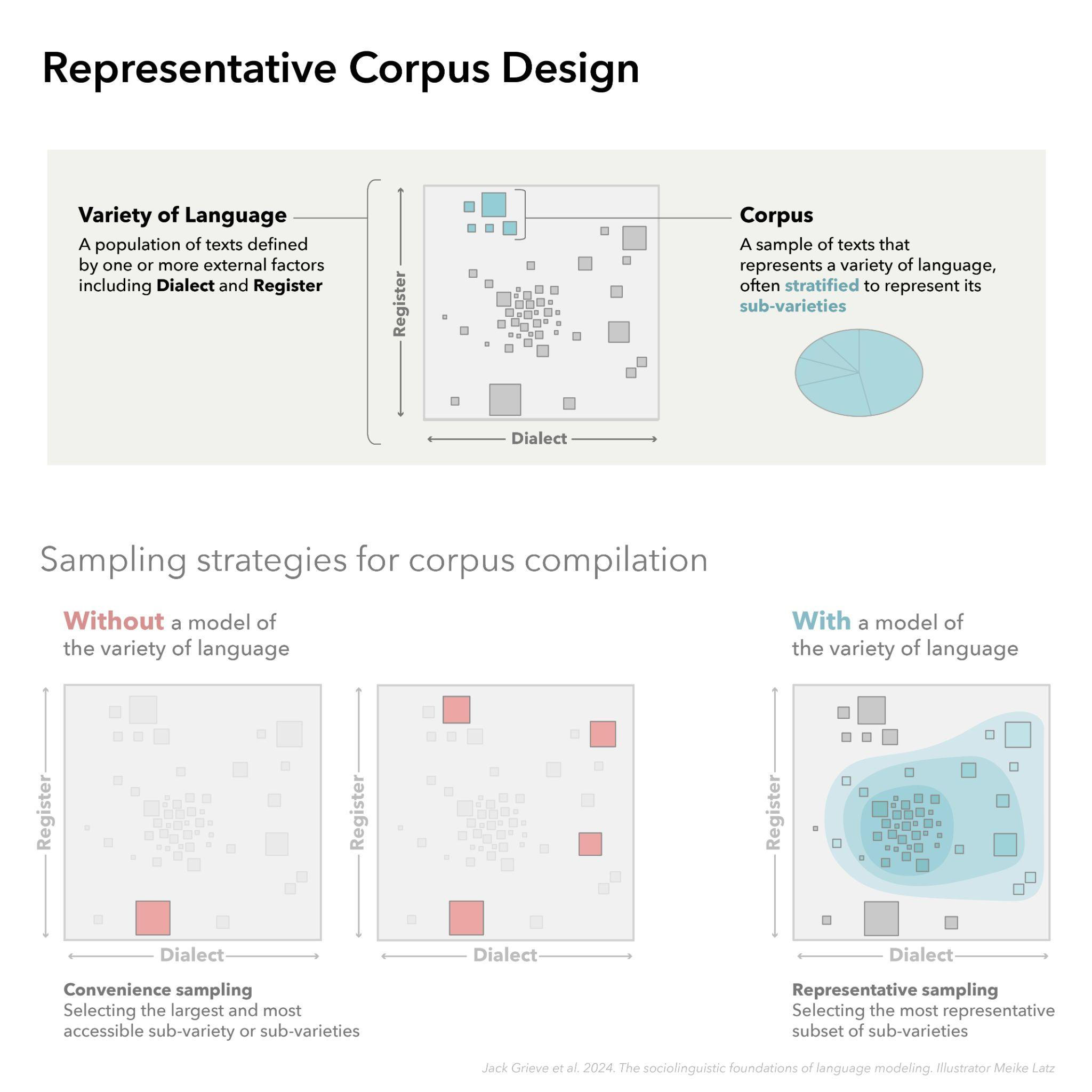}
    \caption{\textit{Representative Corpus Design}. This figure presents a corpus as a representative sample of texts taken from a given variety of language (i.e., from a larger population of texts delimited by relevant extra-linguistic factors). This figure also illustrates how compiling a corpus that accurately represents a target variety requires access to an underlying model of that variety of language, including its internal sub-varieties, so that the corpus can be stratified so as to capture internal variation in that variety. Naïve corpus compilation strategies that rely on convenience sampling will generally lead to less representative samples. 
}
    \label{fig:fig2RepCorpusDesign}
\end{figure}

Notably, this conception of a variety of language is especially common in corpus linguistics, where a corpus is often seen as representing a variety of language: a corpus consists of a sample of texts drawn from the larger population of texts targeted for analysis \cite{Biber1993,McEneryWilson2001,McEneryXiaoTono2006,ScholzPelletierPullumNefdt2024}. The goal of analyzing the structure of language observed in a corpus is therefore to draw generalizations about the variety of language (i.e., the larger population of texts) represented by that corpus. Furthermore, the quality of a corpus, and by extension the generalizability of any analyses based on that corpus, depends directly on the representativeness of this sample, including the accurate identification of its primary constituent sub-varieties (see Figure~\ref{fig:fig2RepCorpusDesign}). 

Finally, if a variety of language is defined as a population of texts delimited by some set of external criteria, the general expectation is that this population of texts will differ from other populations of texts in terms of its linguistic structure, including its grammar, phonology, lexis, and discourse \cite{CrystalDavy1969,Jackson2007}. For example, among other features, a regional dialect may be characterized by the specific pronunciation of certain vowels, whereas a conversational register might be characterized by its rate of use of certain pronouns. Crucially, we can expect that any social group or any social context that is recognized within society will generally become associated with distinct patterns of linguistic variation over time, if only because, at the most basic level, certain words associated with concepts of particular importance to that group or context will be favored or will develop over time, although differences can generally be expected to emerge across all levels of linguistic analysis, depending on the communicative constraints and affordances associated with the extra-linguistic factors that define that variety (for discussion, see \cite{Grieve2023}). Although the number of possible varieties is therefore innumerable, a general goal of linguistic analysis is to identify varieties that are maximally distinctive, for example, mapping the dialect regions of a country \cite{WielingNerbonne2015,Grieve2016}, defining the sub-types of a given register \cite{Biber1989,GrieveBiberFriginalNekrasova2010}, or identifying the most distinct periods of a language \cite{GriesHilpert2008,DegaetanoOrtliebTeich2018}. 

To summarize the discussion presented in this section, we offer the following definition of a variety of language (see Figure~\ref{fig:fig1varietiesoflanguage}):

\begin{quote}
    A \textbf{variety of language} is a population of texts defined by one or more external factors, especially related to the social background of the people who produce these texts, the social context in which these texts are produced, and the period of time over which these texts are produced. 
\end{quote}

Furthermore, we define a \textbf{corpus} as a sample of texts drawn from a specific variety of language, i.e., from a larger population of texts (see Figure~\ref{fig:fig2RepCorpusDesign}). In this sense, we say that a corpus \textit{represents} a given variety of language. It is also important to stress, especially in the context of language modeling, that any corpus – any sample of texts – inherently represents some variety of language, namely, the smallest common variety that encompasses that sample of texts. However, the representativeness of any corpus depends directly on the quality and the size of the sample, as well as the accurate identification of the variety and its sub-varieties from which texts are sampled. For example, a sample consisting of a few conversational transcripts and emails collected in Great Britain could be taken as representing British English, just not very well. 

Our primary contention in this paper is that language models, which are trained on large corpora of natural language, are therefore \textit{inherently} \textit{modeling varieties of language}. In other words, we conceive of language models as models of \textit{language use} – models of how language is used to create texts in the variety of language that the corpus used to train the model represents. Furthermore, like all linguistic models that are based on corpora of natural language, we believe that the validity and value of a language model depends on the degree to which the training corpus accurately represents the variety that is effectively being modeled, which we refer to as the \textbf{target variety} – even if that variety of language is unknown or underspecified. Consequently, our claim is that understanding how to define and represent varieties of language is of direct relevance to language modeling: we believe that many problems that arise in language modeling result from a mismatch between the variety of language that language models are effectively intended to represent and the variety of language that is actually represented by the training corpora. We believe that this perspective is not only novel but fundamental to understanding the nature of language modeling and how to maximize the societal value of LLMs. To support and exemplify this claim, in the remainder of this paper, we therefore consider specific implications of this sociolinguistic conception of language modeling for a range of different challenges currently being faced in language modeling. 

\clearpage
\hspace*{-2in}
\clearpage

\clearpage
\hspace*{-2in}
\begin{figure}
    \centering
    \includegraphics[width=\linewidth]{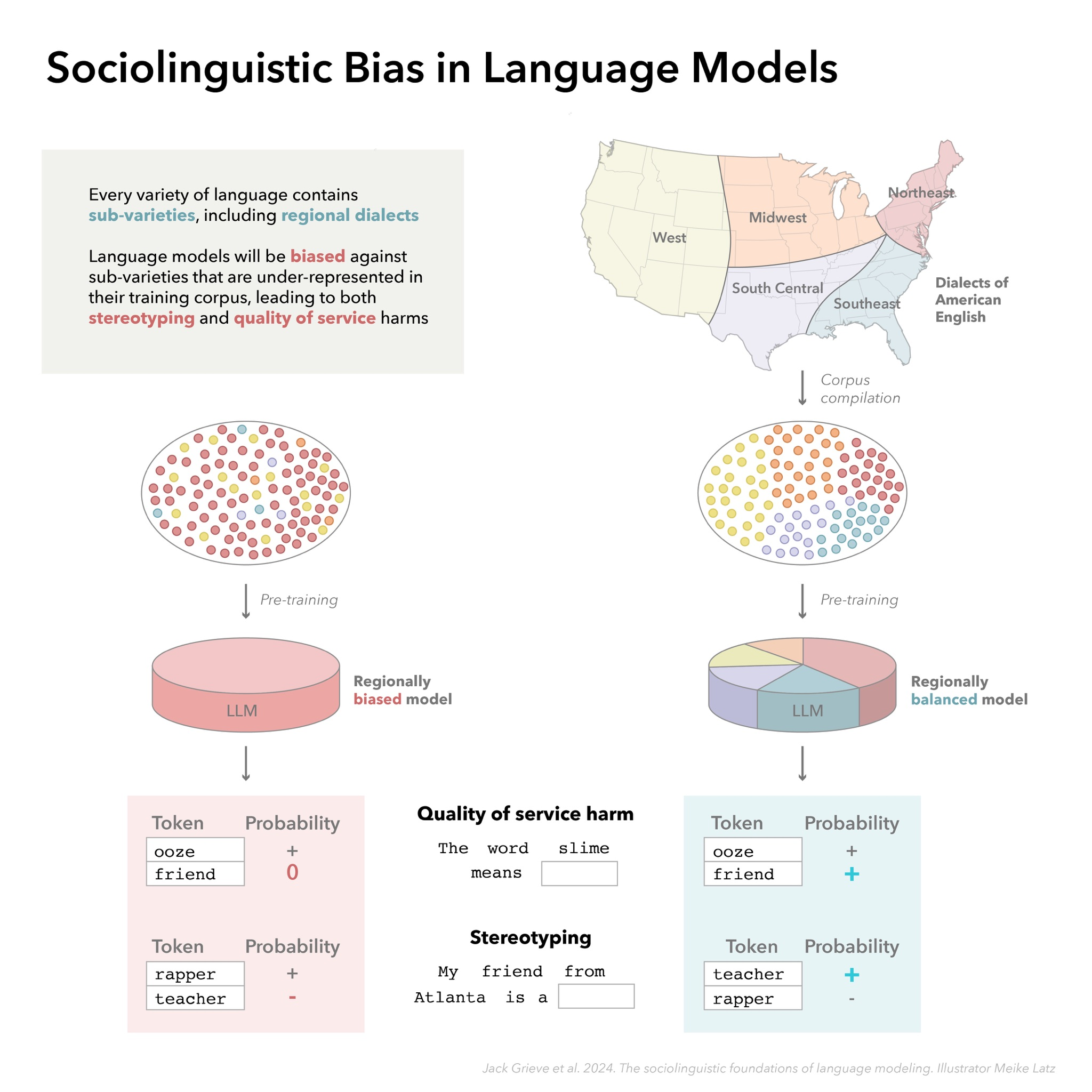}
    \caption{\textit{Sociolinguistic Bias in Language Models}. This figure illustrates how training language models on corpora that accurately represent the target variety of language including its internal structure, especially its constituent dialects, can help address social bias, including both quality-of-service harms and stereotyping. This is exemplified by comparing two hypothetical models which model American English but are trained on corpora that inaccurately and accurately represent regional dialect variation (based on Grieve, 2016) in this larger variety of language.}
    \label{fig:fig3socibiasinlm}
\end{figure}
\clearpage

\section{Challenges}

\subsection{Social Bias}

NLP systems generally suffer from \textit{social bias}: their real-world application leads to outcomes that unfairly disadvantage or harm specific social groups \cite{ShahSchwartzHovy2020,BlodgettBarocasWallach2020,DevShengZhaoAmstutzSunHouChang2022,NavigliConiaRoss2023}. Social bias can be introduced at various points during the development and deployment of NLP systems \cite{HovyPrabhumoye2021}, but given the unsupervised nature of language modeling, training corpora are a key source of social bias in LLMs \cite{BenderGebruMcMillanMajorShmitchell2021,Ferrara2023}. While bias in NLP systems can harm people in various ways \cite{BlodgettBarocasWallach2020}, in this section, we primarily focus on two common harmful outcomes of social bias. These two types of harms are most commonly discussed in terms of \textit{quality-of-service harms} and \textit{stereotyping harms} (e.g., \cite{Crawford2017,Blodgett2021,DevShengZhaoAmstutzSunHouChang2022,Weerts2021}), although many different systems have been proposed for classifying biases and harms in NLP, which define these terms in somewhat different ways, along with many additional and often overlapping categories \cite{BlodgettBarocasWallach2020}. Both of these types of harms are especially relevant to LLMs, and crucially, we believe both can be better understood and addressed in language modeling by adopting a sociolinguistic perspective (see Figure~\ref{fig:fig3socibiasinlm}).

First, social bias can be characterized by poor system performance for certain social groups that are interacting with LLMs and applications based on language models: token prediction will be more or less accurate depending on the social origins of the language inputted into the system. For example, ChatGPT might have difficulty correctly understanding prompts written by people from certain social groups due to their use of non-standard or socially restricted language patterns. This type of bias leads to what is known as \textbf{quality-of-service harms}, where the performance of these systems varies depending on the social background of the user \cite{Crawford2017,DevShengZhaoAmstutzSunHouChang2022}. These types of quality-of-service harms can often be the product of \textbf{selection bias}, as they result from how training data is \textit{selected} from across the society whose language is being modeled \cite{ShahSchwartzHovy2020}: in general, if language data from certain social groups is under-represented in the training data for a language model, we should expect that NLP applications based on that model will process language structures produced by these groups less accurately and consequently exhibit poorer performance for these groups \cite{BlodgettBarocasWallach2020,LahotiBlummMaKotikalapudiPotluriTan2023}. Notably, quality-of-service harms, especially those resulting from selection bias, have been one of the central concerns in computational sociolinguistics \cite{NguyenDogruozRoseDeJong2016,Eisenstein2017,Grieve2023}. Researchers in this emerging field have stressed for the past decade that the performance of NLP systems generally varies for people from different social groups and have called for engagement with description and theory from sociolinguistics to help address this basic form of social bias (e.g. \cite{HovySogaard2015,JorgensenKarrebaekMadsenMoller2015,BlodgettOConnor2017,JurgensTsvetkovJurafsky2017}).

Second, social bias can be characterized by systems that produce outputs that directly harm or discriminate against certain social groups even when they are not directly engaging with these systems themselves. For example, when prompted, ChatGPT might be more likely to produce negative portrayals about ethnicities and genders, no matter who is doing the prompting \cite{BommasaniHudsonAdeliAltmanAroravonArx2021,LahotiBlummMaKotikalapudiPotluriTan2023}. Most notably, this type of bias can lead to what is known as \textbf{stereotyping harms} \cite{Crawford2017}, as well as related harms like \textit{disparagement} and \textit{dehumanization} \cite{DevShengZhaoAmstutzSunHouChang2022}, where negative viewpoints about specific social groups are propagated, as has been widely discussed in regards to LLMs \cite{BenderGebruMcMillanMajorShmitchell2021}. Once again this issue can be traced back to the data the language model was trained on. If the training corpus contains relatively frequent expression of harmful or inaccurate ideas about certain social groups – as we can safely assume any large, unconstrained sample of internet writings will – language models will inevitably reproduce those biases \cite{BenderGebruMcMillanMajorShmitchell2021,Ferrara2023}. As Bender et al. (2021, 613) state, “large, uncurated, Internet-based datasets encode the dominant/hegemonic view, which further harms people at the margins” \cite{BenderGebruMcMillanMajorShmitchell2021}. These types of harms are generally the product of \textbf{semantic bias}, as they result from the meaning relationships between words inferred by the language model based on patterns of co-occurrence observed in the training corpus \cite{ShahSchwartzHovy2020}. 

From a sociolinguistic perspective, we believe that social bias in language modeling can generally be addressed by training on corpora that more accurately represent the target variety of language. It is especially important that the training corpus represents the \textit{internal structure} of the target variety, in the sense that the sub-varieties of that variety of language, including most importantly the major dialects of that variety of language, are adequately represented in the training corpus (see Figure~\ref{fig:fig3socibiasinlm}). For example, a corpus intended to represent American English, but which is primarily composed of texts collected from a specific dialect of American English (e.g., texts written by highly educated, middle-class, white Americans from major coastal cities), cannot adequately represent the full diversity of American English. Any language model trained on such a corpus should therefore be expected to be biased against social groups that are underrepresented in the training data, compared to a language model trained on a corpus that more accurately represents variation in American English. 

The link between corpus design and quality-of-service harms in LLMs is especially clear: because language varies in systematic ways, to ensure a language model can accurately process language \textit{from} a wide range of social groups, it must be trained on corpora that represent the language used \textit{by} a wide range of social groups, i.e., their dialects (see Figure~\ref{fig:fig3socibiasinlm}). For example, consider lexical variation in British and American English: if a model were only trained on American English, it would be much more likely to misinterpret the meaning of words that tend to have different meanings in British English, like \textit{boot }(for \textit{trunk}) or \textit{underground }(for \textit{subway}). Consequently, the quality of service provided by applications based on that model for speakers of British English would be degraded.

Stereotyping and related forms of discrimination generated by LLMs have also often been assumed to result from careless data collection and a lack of data curation \cite{BenderGebruMcMillanMajorShmitchell2021}. A sociolinguistic perspective provides a principled solution to this problem: in general, stereotyping harms can be addressed by using training data that better represents the language produced by a wider range of social groups. One reason that certain social groups are negatively portrayed by LLMs is because they are not allowed to portray themselves in the data used for training. By training on corpora that equitably and deliberately represent the internal varietal structure of the target variety of language, especially the range of dialects of which it is composed, stereotyping and other forms of semantic bias can be mitigated (see Figure~\ref{fig:fig3socibiasinlm}). In other words, modeling data from a wider range of dialects helps ensure that a wider range of viewpoints will be represented by a language model. Stratified corpora that accurately represent the sociolinguistic structure of the target variety can also be used to evaluate and probe a model, allowing for social bias to be identified and interpreted directly.

The sociolinguistic approach to language modeling advocated for in this paper therefore provides a simple yet theoretically grounded basis for understanding the general source of social bias in language modeling, including for addressing both quality-of-service and stereotyping harms, as well as other related types of harms. In addition, a sociolinguistic approach offers a clear pathway for both interpreting and addressing these different forms of social bias during pre-training through careful corpus compilation informed by theories and descriptions of sociolinguistic variation. Crucially, however, such sociolinguistic interventions need not necessarily occur during the initial pre-training of the base model, but can be pursued through the \textit{further pre-training} of base models, as we discuss in the next section.

\subsection{Domain Adaptation}

Despite their remarkable fluency and general applicability, LLMs generally benefit from some form of \textbf{domain} \textbf{adaptation} before deployment \cite{RadfordWuChildLuanAmodeiSutskever2019,GururanganMarasovicSwayamdiptaLoBeltagyDowneySmith2020}. In NLP, domain adaptation is the task of improving the performance of a system that was developed using language data collected in one domain for a different and often more specific domain where the system is to be applied \cite{DaumeIII2007}. Although there are many approaches for adapting language models, including for different downstream tasks (e.g., through forms of supervised and reinforcement learning), in this case, we focus on the process of fine-tuning a base model by extending unsupervised language modeling on a corpus of texts sampled from a specific target \textit{domain} – the real-world context where the system is used, such as texts about a particular topic or from a particular genre \cite{GururanganMarasovicSwayamdiptaLoBeltagyDowneySmith2020,HuWallisAllenZhuLiWangWangChen,HouSalazarPolovets2022}. 

This approach is often referred to as \textbf{further pre-training} because it involves extending the basic form of unsupervised language modeling used to train the base model to new data from the more specific target domain \cite{GururanganMarasovicSwayamdiptaLoBeltagyDowneySmith2020}. The goal is simply to improve the accuracy of token prediction in the target domain, while preserving the underlying fluency of the base model. For example, a base model trained on huge amounts of unrestricted online language data could be adapted to the specific domain of customer service: based on a corpus of customer service transcripts, the parameters of the base model would be adjusted to improve the ability of the model to predict word tokens in texts from that domain given the topics of discussion and the specific types of interactions that characterize that domain.

\begin{figure}
    \centering
    \includegraphics[width=\linewidth]{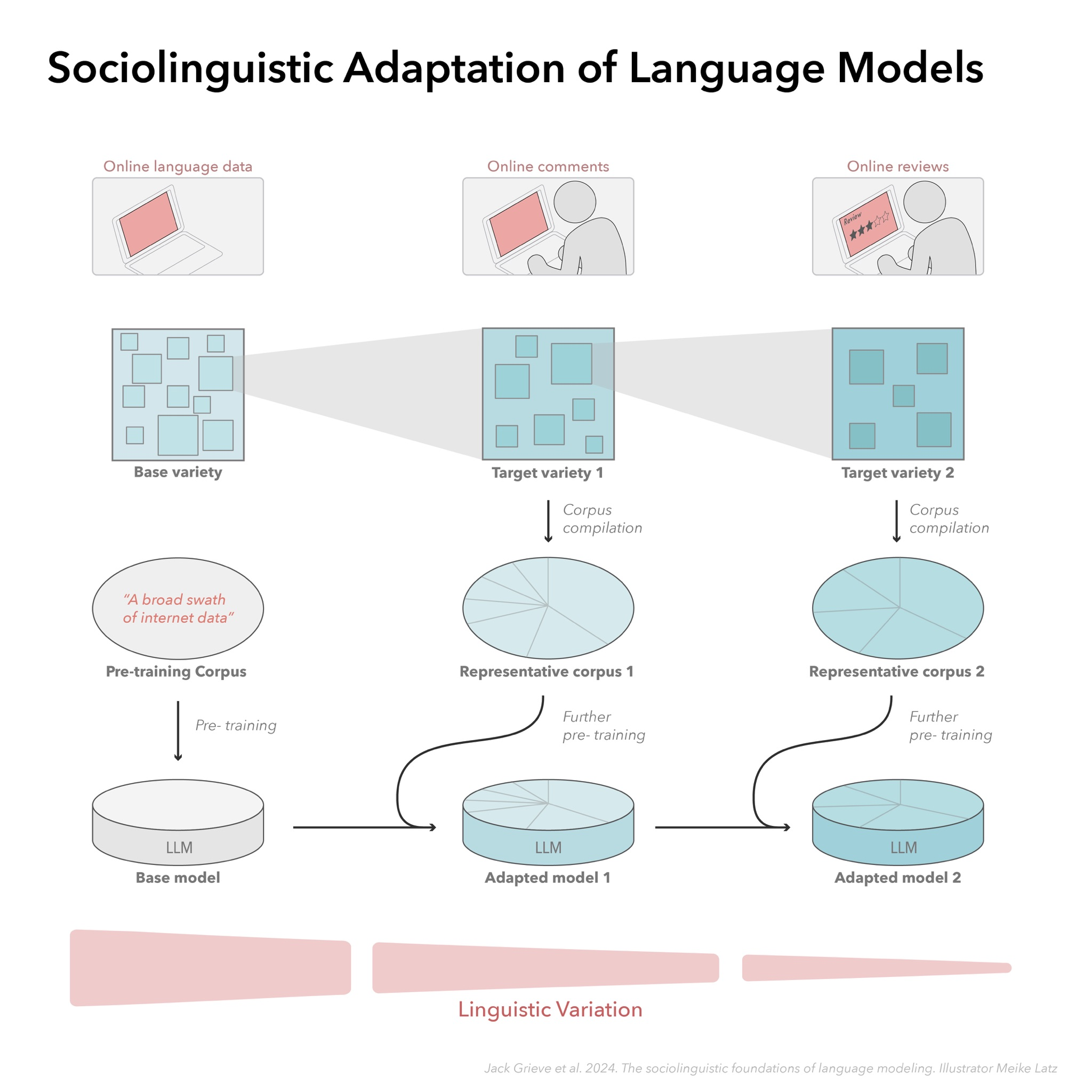}
    \caption{\textit{Sociolinguistic Adaptation of Language Models}. This figure illustrates how an understanding of the sociolinguistic structure of varieties of languages can inform the adaptation of language models. Language model adaptation can be seen as the process of fine-tuning a base model, potentially in an iterative manner, to predict word tokens in a more narrowly defined variety of language that is subsumed by the larger variety of language represented by the base model.}
    \label{fig:fig4socilingadaptationoflm}
\end{figure}

In the context of language modeling, the process of domain adaptation can be straightforwardly reframed directly in sociolinguistic terms (see Figure~\ref{fig:fig4socilingadaptationoflm}). If the goal of the base model is seen as accurately predicting word tokens in a broadly defined variety of language, like the English language, then the goal of domain adaptation can be seen as the process of fine-tuning the base model to allow it to predict word tokens \textit{more accurately} in a more narrowly defined variety of that language – the sub-variety associated with the target domain. Crucially, the adapted model should be expected to be more accurate because more narrowly defined varieties of language \textit{must }be characterized by \textit{less variation} than any larger variety that encompasses it. This process can also potentially be carried out in an iterative manner, where a base model is repeatedly adapted on corpora representing more narrowly defined varieties of language.

A sociolinguistic perspective on domain adaptation therefore sees the target domain as a variety of language, which means that the process of domain adaptation can be meaningfully informed by linguistic analysis that rigorously identifies maximally distinctive varieties of language. This can include both existing research in sociolinguistics, dialectology, and related fields, as well as new research conducted directly to support model training for specified domains. For example, if a base model is adapted for a specific region of the US, research in American dialect geography (e.g., \cite{Grieve2016}) should be consulted to precisely define the sub-region that is being targeted for adaptation (see Figure~\ref{fig:fig3socibiasinlm}). Similarly, if a base model is adapted for a specific type of blog writing, research on register variation in blogs (e.g., \cite{GrieveBiberFriginalNekrasova2010}) should be consulted to precisely define the sub-type of blog writing that is being targeted for adaptation. 

\clearpage
\hspace*{-2in}
\clearpage

Crucially, however, sociolinguistics not only provides a basis for identifying valid targets for domain adaptation but for mapping and modeling the internal structure of these target varieties (see Figure~\ref{fig:fig4socilingadaptationoflm}). This is especially important because target varieties for domain adaptation are often well-defined by default. For example, if a fine-tuning corpus is collected by sampling data from a particular social media platform, a relatively homogeneous variety of language will have naturally been targeted; however, a random sample of texts from that variety, drawn without taking into account its internal structure, might severely under-represent sub-varieties of interest (see Figure~\ref{fig:fig2RepCorpusDesign}). For example, a social media corpus may be dominated by certain sub-registers (e.g., abusive or promotional posts) that are not the target of adaptation, while the sub-registers that are the target of adaptation (e.g., interactive or informational posts) may be limited. Similarly, people from certain social groups may be underrepresented in specific domains, resulting in social bias being inadvertently exacerbated by naive domain adaptation. In many cases, the target variety cannot even be accurately defined until the overall structure of the larger variety in which it is subsumed is understood through careful sociolinguistic analysis.

A sociolinguistic perspective also highlights a more general problem with domain adaptation: the success of this process depends on the relationship between the larger variety represented by the base model and the smaller target variety towards which the base model is being adapted. Ideally the variety of language represented by the base model would completely subsume the target variety: the target variety would be a sub-variety of the base variety, regardless of whether it was represented directly in the base training data. However, the target variety may not be adequately represented in the data sampled for training the base model. For example, the target variety could be associated with a social group or a social context that is severely underrepresented in the base training corpus. In such situations, fine-tuning regimes informed by sociolinguistic theory and description would likely be beneficial.

Finally, understanding the sociolinguistic structure of the larger variety of language could also allow models to be adapted to represent target varieties with missing data. For example, if empirical research in linguistics has found that a target dialect or register for which data is lacking falls between multiple dialects or registers for which data is available, a model could be adapted for the target variety by training on a combination of the available corpora. Overlap between varieties could also be exploited in a similar way: for example, if data is lacking for a target variety defined in terms of a specific register and a specific dialect, a model could be adapted for the target variety by fine-tuning on a combination of corpora that represent that specific dialect and that specific register. These types of techniques could even be used to create a model of a variety of language that does not yet exist – engineered by training on corpora representing different registers and dialects.

\subsection{Alignment}

The related challenges of social bias and domain adaptation can be seen as forms of the more general \textit{alignment} \textit{problem} – how to ensure that the behavior of AI systems aligns with the values and expectations of society \cite{Gabriel2020,HendrycksBurnsBasartCritchLiSongSteinhardt2020,Christian2021,NgoChanMindermann2022,Dung2023}. Misalignment arises not simply when AI systems fail to achieve their intended goals, but when they pursue these goals, even successfully, in ways that have negative or unforeseen consequences or that are not in accordance with societal values, for example, in ways society finds to be inappropriate, unethical, immoral, or dishonest. \textbf{Alignment} is therefore the general process of guiding AI systems to behave in ways that are consistent with the broader expectations of society, while discouraging them from behaving in ways that are inconsistent with these expectations, especially to avoid unintended risks and harms. Crucially, the challenge is not only \textit{how} to guide AI systems but \textit{where} to guide them \cite{Gabriel2020}.

Although alignment is a long-standing concern in AI \cite{Wiener1960}, attention has grown in recent years due to the growing complexity and ubiquity of real-world AI systems, especially systems based on language modeling \cite{ShenJinHuangLiuDongGuo2023,LiuZhangFengVosoughi2022,LiuYangJiaZhangZhouDai2023,WangZhongLiMiZengHuang2023,WolfWiesLevineShashua2023}, which potentially allow for misalignment to emerge on many different levels \cite{Gabriel2020,Dung2023}. For example, consider a generative language model that automatically produces reviews of scientific literature on a specified topic. An obviously misaligned system might produce reviews that are clearly wrong – incoherent or incorrect – while a less obviously misaligned system might produce fluent reviews, completing the task successfully in a superficial way, but getting facts wrong, for example, referencing publications that do not exist. This type of a \textit{hallucination} – the presentation of false information as if it is true – is a common form of misalignment in LLMs \cite{EvansCottonBarrattFinnvedenBalesBalwitWills2021,TonmoyZamanJainRaniRawteChadhaDas2024}. A more insidiously misaligned system, however, might produce perfectly accurate and fluent syntheses that cite relevant literature, but exhibit other problematic behaviors, such as limiting references to certain ideas or researchers in certain fields, thereby effectively suppressing certain viewpoints \cite{BenderGebruMcMillanMajorShmitchell2021}.

One solution for aligning language models with the values of society is by training these models using corpora that are in some way deemed to be more aligned with these values (e.g., \cite{SolaimanDennison2021}. As we have argued throughout this paper, we believe sociolinguistic theory provides a meaningful, interpretable, and productive way to guide this process – allowing us to better understand how corpora can be compiled so as to allow for societal expectations to be captured, crucially without pre-specifying what these exact expectations are. 

Our view is that alignment is possible if training corpora accurately represent the range of dialects and registers of the target variety. In terms of dialects, as we discussed when we considered the challenge of social bias, we believe that, by balancing training data originating from different social groups, language models can be trained to better align with the general values of society, as opposed to the values of some particular social group. Similarly, in terms of registers, as we discussed when we considered the challenge of domain adaptation, we believe that, by balancing training data originating from different communicative contexts, language models can be trained to better align with the expectation that they will perform adequately across the range of contexts found in that society. In other words, the values and expectations of a society are instantiated in their patterns of language use. In general, we therefore believe that a major source of LLM misalignment results from what we call \textbf{varietal misalignment} and that LLM misalignment can therefore be addressed, at least in part, by aligning training corpora to the varietal structure of the target variety. 

In addition to addressing alignment issues related to social bias and domain adaptation, we believe this sociolinguistic approach can potentially help us train models that are less susceptible to unethical and dishonest behavior in general, because respecting sociolinguistic diversity entails training models on data that represents a greater diversity of viewpoints, experiences, and contexts. As LLMs are models of varieties of language, they will be better models, more aligned with the needs, expectations, and values of society, when they account for the full range of sub-varieties, and hence the full range of perspectives, found within that society.

Finally, it is important to acknowledge that while a sociolinguistic perspective provides a basis for aligning a language model to the general viewpoints of the society that it is intended to serve, this approach does not ensure that the resultant language model will be aligned with the ethical and moral \textit{aspirations} of that society. For example, a generative language model trained on a socially balanced corpus of the English language will still potentially produce texts that express racist viewpoints because a portion of English texts expresses racist viewpoints. There might be greater equity in the types of stereotypes it spreads, but such behavior can still be seen as a form of misalignment. A sociolinguistic perspective, however, also provides a possible solution to this problem – by deliberately weighting the varieties of language represented in the training corpus. For example, if a particular social group has been broadly disadvantaged or has a worldview that society wishes to encourage, the portion of the corpus representing the relevant varieties of language can be more heavily weighted during training. In this way, a sociolinguistic perspective can provide a theoretical basis not only for \textit{balancing }but for \textit{controlling} the alignment of language models.

\subsection{Language Change}

Thus far, our discussion has focused on how a series of challenges in language modeling related to bias, adaptation, and alignment can be addressed, in principle, by building training corpora that better represent the dialects and registers of the target variety. Another form of this basic problem involves ensuring that language models and applications based on language models are responsive to language change and cultural change more generally \cite{BenderGebruMcMillanMajorShmitchell2021,BommasaniHudsonAdeliAltmanAroravonArx2021}. All varieties of language change over time, often in ways that are difficult, if not impossible, to predict \cite{Lass1997}. If language models are to maintain their fluency and not become obsolete, they must therefore be continuously updated using training corpora that consist of examples of contemporary language use. In principle, this problem can be resolved by compiling new corpora over time that \textit{consistently} represent the target variety and its evolving internal varietal structure. The challenge is therefore to understand how the sociolinguistic landscape of registers and dialects of that variety of language has changed over time, which can only be accomplished accurately through detailed and ongoing sociolinguistic analysis. 

A related issue that has caused growing concern in language modeling is that over time more and more real-world language will presumably be produced with the assistance of LLMs, which will make it increasingly difficult to compile contemporary corpora of \textit{real} human language for training new models or updating existing ones \cite{ShumailovShumaylovZhaoGalPapernotAnderson2023}. Proposed solutions to these problems of \textit{data contamination} \cite{BalloccuSchmidtovaLangoDuvsek2024} and \textit{task contamination} \cite{LiFlanigan2024} generally involve finding ways to exclude machine-generated language from future training data, including through watermarking systems \cite{KirchenbauerGeipingWenKatzMiersGoldstein2023}. These types of solutions, however, seem easy to confound, if only because they do not generally allow texts written collaboratively by human and machine to be identified, which is likely to become increasingly common and diversified in everyday life. 

Despite real concerns about LLM detection in certain contexts, the rising use of LLMs to generate language is not difficult to reconcile with sociolinguistic theory and practice. Over time, AI systems based on language models will undoubtedly start to change how we use language. Texts generated with the help of language models will increasingly enter into the real world. At this point, from an externalist perspective \cite{ScholzPelletierPullumNefdt2024}, these texts will be part of language – produced, transmitted, and understood by humans as language, often indistinguishable from human-generated language in the regular flow of real-world language use. Ultimately, the distinction between human- and machine-generated language can therefore be seen as simply another aspect of register that defines variation within varieties of language, just like all communicative technologies that have come before, including the invention of writing and digital communication. 

Taking a sociolinguistic perspective, it is also important to acknowledge that the rise of language models is creating \textit{new} varieties of language, including those characterized by the linguistic interaction between humans and machines, such as dialogues with ChatGPT. These new varieties, which will only continue to diversify over time, will also need to be accounted for, like all varieties of language, both by theories of sociolinguistic variation and by the evolving language models designed to represent contemporary language use. If language models are to be kept up-to-date, machine-generated language cannot be excluded, as its production will become a significant driver of language change.

\subsection{Scale}

In addition to more specific insights into the development and deployment of language models, we believe a sociolinguistic perspective can also help to explain the remarkable success of LLMs more generally, which has been attributed both to the development of new deep learning architectures and the use of extremely large corpora of natural language for training \cite{KaplanMcCandlishHenighanBrownChessChild2020,BenderGebruMcMillanMajorShmitchell2021,BommasaniHudsonAdeliAltmanAroravonArx2021}. Although there is a clear relationship between the scale of the training data and the success of these systems, it is not altogether clear \textit{why} increasing the amount of training data results in such great increases in performance. Is there a limit to how much performance can be gained simply by increasing the scale of the training data? How can more powerful models be developed with less data? These are fundamental questions for LLM development \cite{BommasaniHudsonAdeliAltmanAroravonArx2021}, especially because of the significant costs and environmental impacts associated with increases in scale \cite{BenderGebruMcMillanMajorShmitchell2021}. We believe these are questions that can be uniquely informed by a sociolinguistic perspective.

The obvious reason why increasing the amount of training data provided to the model improves the performance of a language model is that it provides access to a wider range of language patterns. Working with extremely large corpora is clearly necessary – the complexity of language demands it – but it is also clear that scale is not sufficient on its own. For example, a language model trained repeatedly on the same dataset will not improve. What therefore matters is not simply the \textit{scale} of the training data but the \textit{diversity} of the training data. 

Although the importance of the diversity of training data has often been stressed in critiques of LLMs \cite{BrownMannRyderSubbiahKaplanDhariwal2020,BenderGebruMcMillanMajorShmitchell2021}, the sociolinguistic perspective advocated in this paper provides a theoretical basis for understanding this relationship with greater precision: diversity in the training corpus, in terms of both its linguistic structure and its semantic content, can be seen as directly reflecting the diversity of the varieties of language represented by that corpus. To maximize the performance of language models and the efficiency with which these improvements can be obtained, in our view, it is therefore far more important to focus on increasing the varietal diversity of the training data than purely its scale. This can be achieved by carefully representing a wider range of contemporary language varieties in the training corpora, including both dialects and registers, as we have discussed throughout this paper.  

This sociolinguistic perspective also provides an answer to questions about the limits of increasing the scale of training data \cite{BommasaniHudsonAdeliAltmanAroravonArx2021}. At what point should increasing the size of the training corpus no longer lead to substantial improvements in model performance? Our hypothesis is that increasing the scale of training data will continue to increase the performance of language models so long as it also results in an increase in the sociolinguistic diversity in the training corpus. Crucially, this implies that attempts to empirically assess the limits of scale simply by comparing model performance as the amount of training data increases will not be accurate, unless the sociolinguistic diversity of the corpus is also controlled for and measured alongside corpus size. 

Finally, a sociolinguistic perspective also offers clear direction for training models using limited amounts of data, for example, for under-resourced languages \cite{BenderGebruMcMillanMajorShmitchell2021,RameshSitaramChoudhury2023}: models can be developed on a smaller scale by taking care to maximize the amount of sociolinguistic diversity in the training data, given the target variety. Moving forward, we therefore believe that optimizing the development and performance of LLMs will necessarily involve incorporating insights from sociolinguistics to enhance the diversity and representativeness of language data used for training.

\section{Conclusion}

In this paper, we have advanced the claim that language models \textit{inherently} represent varieties of language. By extension, we have also argued that the performance, utility, and ethical application of language models depends directly on how well training corpora represent the varieties of language being modeled, including their internal varietal structure. Our view is that the societal value of language models in general depends not only on the amount of language data used for training but on the sociolinguistic diversity and representativeness of these corpora. We therefore believe that incorporating insights from sociolinguistics is crucial to the future of language modeling. To support this claim, we have identified several ways in which a sociolinguistic perspective can provide a basis for addressing specific challenges in language modeling related to social bias, domain adaptation, alignment, language change, and scale in a principled and unified manner.

Notably, there already has been considerable discussion of these types of challenges in language modeling and NLP more generally, with proposals to address these issues often emphasizing the need for more careful curation of training data \cite{BenderGebruMcMillanMajorShmitchell2021,HovyPrabhumoye2021} and for incorporating social and even sociolinguistic insight into these models \cite{Hovy2018,HovyYang2021,NguyenRosseelGrieve2021,YangHovyJurgensPlank2024}, especially within the emerging field of computational sociolinguistics \cite{NguyenDogruozRoseDeJong2016,Grieve2023}. For example, to address risks related to social bias in LLMs, Bender et al. (2021, 610) recommend that resources must be invested for “curating and carefully documenting datasets rather than ingesting everything on the web” \cite{BenderGebruMcMillanMajorShmitchell2021}, while Yang et al. (2024, 1) argue that issues with LLM performance are related to “a lack of awareness of the factors, context, and implications of the social environment in which NLP operates, which we call \textit{social awareness}” \cite{YangHovyJurgensPlank2024}. 

What is lacking in these discussions, however, is the proposal of a general linguistic framework for solving these types of problems within the basic paradigm of language modeling, especially one that is theoretically grounded in our scientific understanding of language variation and change. Although the lack of social diversity in training data has been repeatedly identified as a problem for LLMs, what exactly this means and how exactly this can be measured and addressed in a principled manner has not been articulated. 

Given this emerging discourse, the primary contribution of this paper is to propose a theoretical and empirical foundation for addressing a wide range of challenges in language modeling that is based directly on sociolinguistic theory, specifically the concept of a \textit{variety of language} – a topic that to the best of our knowledge has been absent from discussions of language modeling up until now, even within computational sociolinguistics. This perspective is also notably quite different from discussions of language modeling in linguistics, which have focused on the status of LLMs as \textit{models of language cognition} \cite{Piantadosi2023,DentellaGuntherLeivada2023,MarcusLeivadaMurphy2023}. In this paper, we have attempted to shift this discussion, focusing instead on understanding language models as \textit{models of language use}, which we believe has far more direct and immediate consequences for the development and deployment of language models in the real world. 

Our basic claim is therefore that language models can be improved in many ways by training on datasets that endeavor to accurately represent the varieties of language being modeled. We therefore believe that there is a clear and urgent need for sociolinguistic insight in language model design and evaluation. At the most basic level, language models are models of how language is used for communication within society. Understanding the structure of society, and how this structure is reflected in patterns of language use, is therefore critical to maximizing the benefits of language models for the societies in which they are increasingly being embedded. Moving forward, we believe that research on language use – not only in sociolinguistics, but in corpus linguistics, discourse analysis, pragmatics, cognitive linguistics, and other fields of linguistics that focus on understanding how language is used for communication in the real world – will increasingly become central to advancing the field of language modeling, as well as NLP and AI more generally. 

\section*{Acknowledgement}

We would especially like to thank Dong Nguyen for her comments on this paper, as well as Meike Latz for creating the artwork presented in this paper. This paper also benefited from discussions with Su Lin Blodgett, Dirk Hovy, Huang He, David Jurgens, Taylor Jones, and Emily Waibel. Sara Bartl, Alejandro Jawerbaum, and Dana Roemling were supported by the UKRI ESRC Midlands Graduate School Doctoral Training Partnership ES/P000711/1. Bodo Winter was supported by the UKRI Future Leaders Fellowship MR/T040505/1.

\clearpage
\bibliographystyle{unsrt}  
\bibliography{references}

\end{document}